\documentclass[letterpaper, 10 pt, conference]{ieeeconf}  

\IEEEoverridecommandlockouts                              

\usepackage[left=1.88cm,right=1.88cm,top=1.88cm,bottom=1.88cm]{geometry}
\usepackage{amsmath,amssymb,amsfonts}
\usepackage{cite}
\usepackage{textcomp}
\usepackage{xcolor}
\let\labelindent\relax
\usepackage{enumitem}
\usepackage{tabularx}
\usepackage{soul}
\usepackage{times}
\usepackage{multicol}
\usepackage{bm}
\usepackage[nolist]{acronym}

\usepackage[linesnumbered,ruled]{algorithm2e}
\usepackage{listings}
\usepackage{mathtools}
\usepackage{array}
\usepackage{diagbox}
\usepackage{etoolbox}\AtBeginEnvironment{algorithmic}{\small}
\usepackage{algorithmic}

\usepackage{lipsum}
\usepackage[colorinlistoftodos]{todonotes}

\usepackage{graphicx}
\usepackage{caption}
\usepackage{subcaption}

\newcommand{\blue}[1]{\protect\color{blue}{#1}\protect\color{black}}

\usepackage{amsthm}

\usepackage{bbm} 
\usepackage[colorlinks = True, linkcolor = blue, citecolor = blue]{hyperref}

\title{\LARGE \bf
KGLAMP: Knowledge Graph-guided Language model for Adaptive Multi-robot Planning and Replanning
}

\author{
Chak Lam Shek$^{1,2}$ \quad  Faizan M. Tariq$^1$  \quad Sangjae Bae$^1$ \quad David Isele$^1$ \quad Piyush Gupta $^1$ 
\thanks{$^1$Honda Research Institute, USA, San Jose, CA, 95134.}
\thanks{$^2$University of Maryland, College Park, MD 20742, USA.}
\thanks{All work was performed during the internship of Chak Lam Shek while he was employed at HRI.} 
\thanks{Corresponding author: \href{mailto:piyush_gupta@honda-ri.com}{\texttt{piyush\_gupta@honda-ri.com}}.}
\thanks{This work has been submitted to the IEEE for possible publication. Copyright may be transferred without notice, after which this version may no longer be accessible.}
}

\begin{document}

\maketitle


\begin{abstract}
Heterogeneous multi-robot systems are increasingly used in long-horizon missions requiring coordinated planning across diverse capabilities. However, existing planning approaches struggle to construct accurate symbolic representations and maintain plan consistency in dynamic environments. Classical PDDL planners require manually crafted symbolic models, while LLM-based planners often ignore agent heterogeneity and environmental uncertainty. We introduce \textit{KGLAMP}, a knowledge-graph–guided LLM planning framework for heterogeneous multi-robot teams. The framework maintains a structured knowledge graph encoding object relations, spatial reachability, and robot capabilities, which guides the LLM in generating accurate PDDL problem specifications. The knowledge graph serves as a persistent, dynamically updated memory that incorporates new observations and triggers replanning upon detecting inconsistencies, enabling symbolic plans to adapt to evolving world states. Experiments on the MAT-THOR benchmark show that KGLAMP improves performance by at least 25.3\% over both LLM-only and PDDL-based variants.


\end{abstract}

\section{INTRODUCTION}

Heterogeneous multi-robot systems are increasingly used in real-world missions such as disaster response, warehouse logistics, and large-scale inspection~\cite{zhang2024multi}. By combining robots with diverse capabilities, they can tackle tasks beyond the reach of homogeneous teams. However, this heterogeneity introduces significant challenges in long-horizon planning, including task allocation and coordinated action in dynamic, uncertain environments~\cite{bai2021multi}. Recent work has therefore focused on planning strategies that better integrate robots with diverse roles and capabilities.

Despite recent progress, effective planning for heterogeneous teams remains challenging~\cite{gupta2022incentivizing, gupta2019achieving}. Planning Domain Definition Language (PDDL)-based approaches offer strong guarantees for long-horizon planning but depend on manually crafted domain and problem specifications~\cite{liu2024autonomous}. These specifications are labor-intensive to construct, and even minor modeling errors can induce brittle behavior, as planners assume complete and consistent environmental representations.

\begin{figure}[ht]
\centering
\begin{subfigure}[t]{0.23\textwidth}
\centering
\includegraphics[width=1.0\linewidth, height=1.0\linewidth, keepaspectratio]{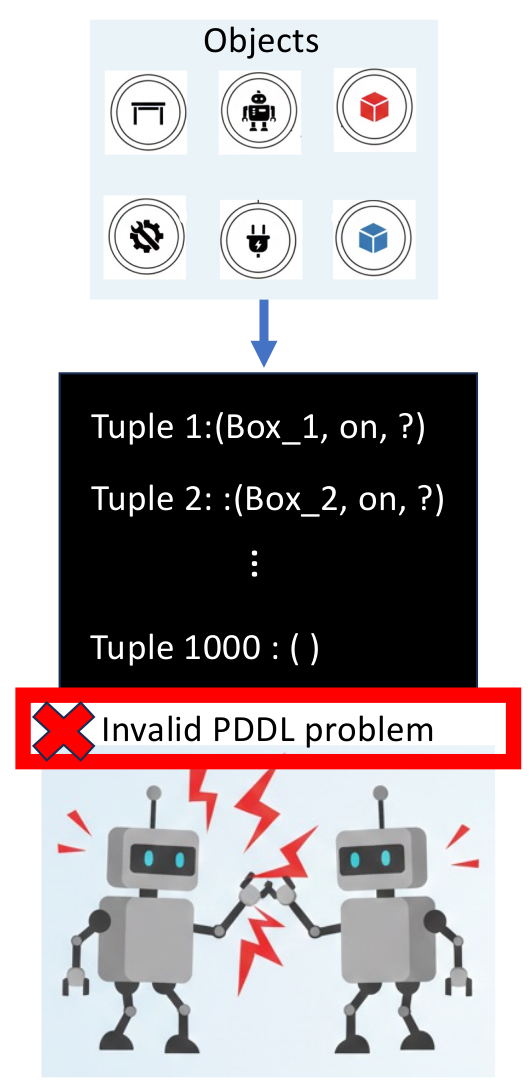}
\caption{Without relational knowledge}
\label{fig:intro1}
\end{subfigure}
~~
\begin{subfigure}[t]{0.23\textwidth}
\centering
\includegraphics[width=1.0\linewidth, height=1.0\linewidth, keepaspectratio]{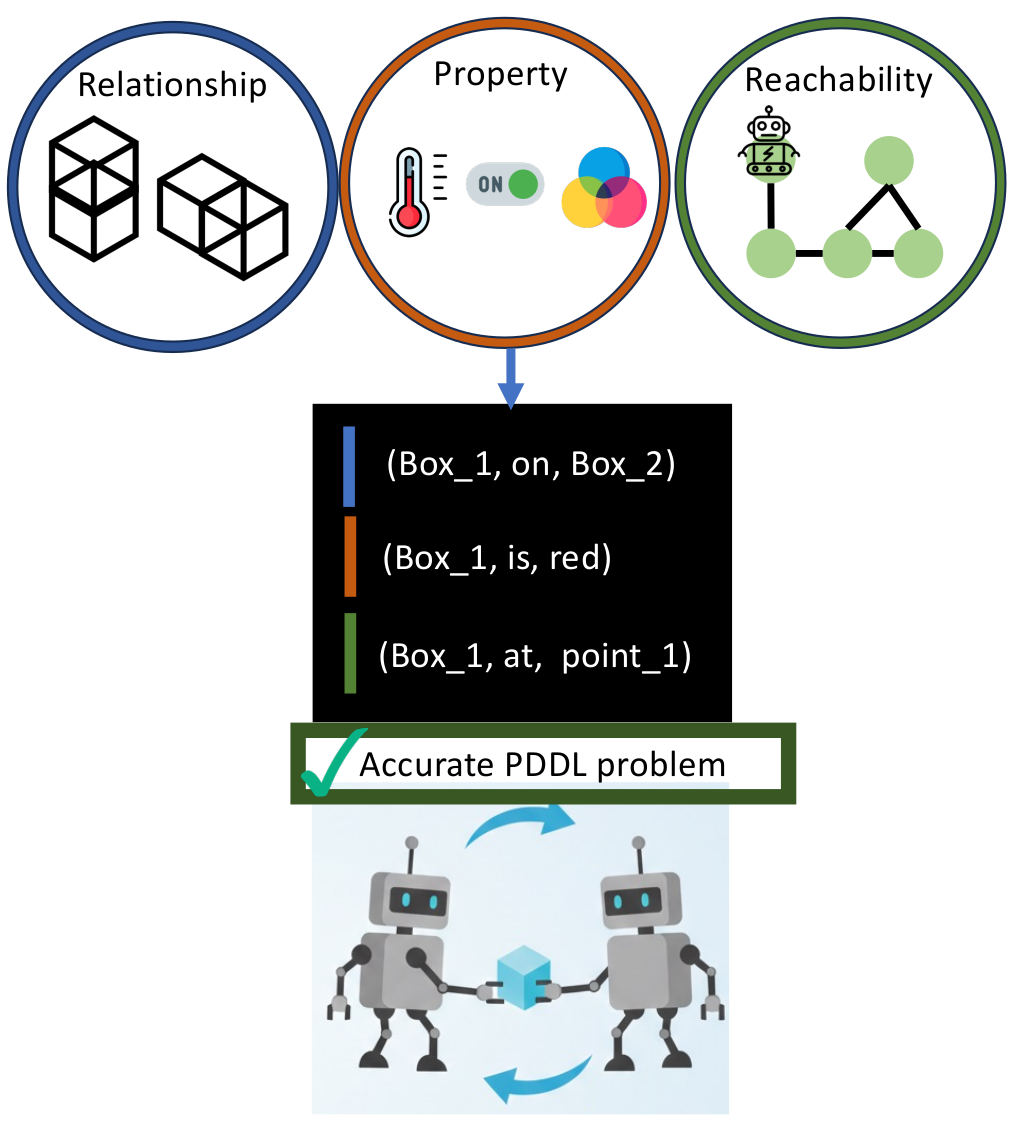}
\caption{With appropriate knowledge graphs}
\label{fig:intro2}
\end{subfigure}
\caption{\small Impact of relational knowledge on task planning. (a) Without relational graphs, PDDL models miss object relationships, leading to failed plans. (b) Incorporating relationship, property, and reachability graphs enables accurate PDDL generation and feasible plans.}
\label{fig:Concept}
\end{figure}
Recent advances in large language models (LLMs) reduce the need for manual domain specification~\cite{liu2023llm+} and enable flexible high-level reasoning for long-horizon planning~\cite{ren2023robots}. Prior work shows that LLMs can generate PDDL and support symbolic planners~\cite{guan2023leveraging}. However, applying LLMs to heterogeneous multi-robot systems remains challenging~\cite{ferreira2024distributed}, as most approaches assume shared action models and identical capabilities, limiting reasoning over embodiment, skills, and task feasibility~\cite{zhang2025lamma}. As illustrated in Fig.~\ref{fig:Concept}, lacking relational context often leads to plans that miss inter-robot dependencies, coordination, and shared resource constraints essential for effective cooperation.

Another limitation is the assumption of complete and accurate environment knowledge. In real-world settings, object states, spatial layouts, and task-relevant properties are often inconsistent or incomplete~\cite{gong2025zero}, causing mismatches between symbolic representations and actual conditions. Effective systems must therefore update their knowledge and adapt plans as new information arrives, which is crucial for reliable long-horizon planning in heterogeneous teams.

To address these challenges, we propose \textit{KGLAMP}, a knowledge-graph–guided LLM planning framework for heterogeneous multi-robot systems. KGLAMP represents environment structure and robot capabilities as a structured knowledge graph capturing object relations, spatial connectivity, and agent-specific properties (Fig.~\ref{fig:intro2}). This representation guides the LLM to generate more accurate planning domain and problem descriptions.

The knowledge graph also acts as a persistent, updatable memory, integrating new observations, resolving inconsistencies, and triggering replanning when needed. This enables robust adaptation to evolving environments and supports reliable coordination in heterogeneous multi-robot teams.

The main contributions of this work are:
\begin{enumerate}
\item A unified \textit{knowledge-graph representation that grounds LLM-based planning} by modeling object relations, spatial reachability, and semantic properties.
\item An incremental knowledge-graph update mechanism that integrates new observations, maintains consistency, and enables \textit{replanning in dynamic environments}.
\item An empirical evaluation on heterogeneous multi-robot tasks showing improved robustness and coordination over LLM-only and PDDL-based baselines.
\end{enumerate}


\section{Related work}\label{Sec: related_work}

Recent work has extended long-horizon planning to more realistic settings~\cite{goel2024novelgym}. However, traditional symbolic planners like PDDL struggle with real-world complexity and uncertainty, motivating LLM-based approaches~\cite{kang2025gflowvlm}.

Structured representations improve interpretability and robustness in complex tasks. Hierarchical methods such as Planning with Hierarchical Trees~\cite{11128711} and Behavior Trees~\cite{cao2023robot} enable scalable behavior through modular decomposition. In open-world settings, skill graphs (e.g., Plan4MC~\cite{yuan2023skill}) and iterative frameworks like PLAN-AND-ACT~\cite{erdogan2025plan} further enhance long-horizon reliability. These ideas extend to multi-agent settings~\cite{bai2021multi}, where coordination is critical. Approaches like SMART-LLM~\cite{kannan2024smart} and graph-enhanced LLMs~\cite{gupta2025graph, lin2024graph} model dependencies to improve coordination.

Recent work has moved beyond homogeneous teams to heterogeneous systems with diverse capabilities, introducing challenges in competency-aware task allocation, cross-skill coordination, and dynamic role adaptation. Approaches such as compositional coordination~\cite{huang2025compositional}, LLM-guided collaboration~\cite{liu2025coherent}, and LLM-driven PDDL planning~\cite{zhang2025lamma} address these issues.
Despite strong benchmark results, many methods struggle to generalize due to inconsistent reasoning and the lack of persistent, task-relevant state—issues amplified in heterogeneous systems requiring shared understanding. While prior work explores memory mechanisms for long-horizon reasoning—including multi-memory architectures~\cite{yanfangzhou2025m2pa}, 
retrieval-augmented planning~\cite{kagaya2024rap}, spatio-temporal navigation memory~\cite{anwar2025remembr}, hybrid multimodal memory~\cite{li2024optimus}, long and short term memory systems~\cite{wang2025karma}, and lifelong planning memory~\cite{agarwal2025l3m+}—most lack structured representations for tracking evolving goals, heterogeneous agent states, and environmental changes across planning iterations. Motivated by these limitations, we introduce a memory-management framework for heterogeneous multi-agent planning that enables persistent, structured state retention and improves coordination among robots.


\section{Problem Formulation and Preliminaries}\label{Sec: Problem_Formulation}

We study long-horizon planning in heterogeneous multi-robot systems with diverse skills and constraints. For example, in a household setting, robots collaborate to prepare food from high-level natural language instructions that omit explicit actions and constraints. Solving such tasks requires inferring task structure, decomposing it into sub-tasks, reasoning about dependencies, and allocating responsibilities across robots that can act in parallel when possible.

\subsection{Multi-Agent Planning (MAP) Formulation}
We formulate the problem as a cooperative Multi-Agent Planning (MAP) framework, defined by
$\langle \mathbb{R}, \mathbb{D}, \{\mathbb{A}_i\}_{i=1}^n, \mathbb{P}, I, G \rangle$,
where $\mathbb{R}=\{r_1,\ldots,r_n\}$ is the set of robots. Each robot $r_i$ has a domain $d_i \in \mathbb{D}$ capturing its capabilities and constraints, and an action set $\mathbb{A}_i$ defining its state transitions. The environment state is represented by logical predicates $\mathbb{P}$, with initial state $I \subseteq \mathbb{P}$ and goal conditions $G \subseteq \mathbb{P}$.

A plan is defined as $\Pi=(\Delta,\prec)$, where $\Delta$ is a set of instantiated robot actions and $\prec$ is a partial order capturing causal and temporal constraints. The plan is valid if executing it from $I$ reaches a state satisfying $G$.

\subsection{Planning Domain Definition Language (PDDL)}
 \begin{figure}[t]
 \centering
 \begin{subfigure}[t]{0.35\textwidth}
 \centering
 \includegraphics[width=1\linewidth, height=1.0\linewidth, keepaspectratio]{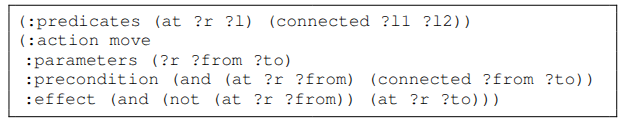}
 \caption{Domain PDDL (STRIPS action schema)}
 \label{fig:pddl-domain}
 \end{subfigure}
 \hfill
 \begin{subfigure}[t]{0.35\textwidth}
 \centering
 \includegraphics[width=1\linewidth, height=1.0\linewidth, keepaspectratio]{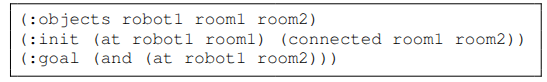}
 \caption{Problem PDDL (task instance)}
 \label{fig:pddl-problem}
 \end{subfigure}

 \caption{\small Minimal STRIPS PDDL example illustrating (a) Domain PDDL and (b) Problem PDDL }
 \label{fig:pddl-example}
 \end{figure}
In PDDL, a planning task is specified by a domain file (Fig.~\ref{fig:pddl-domain}) and a problem file (Fig.~\ref{fig:pddl-problem}). The domain encodes reusable knowledge, including predicates and parameterized action schemas with preconditions and effects defining deterministic state transitions. We focus on STRIPS-style planning~\cite{aineto2018learning}, where preconditions are conjunctions of positive literals and effects are add–delete lists.

The problem file defines a specific instance with a set of objects, an initial state $I$ (under the closed-world assumption), and a goal condition $G$ as a conjunction of ground literals. A planner then computes a sequence of grounded actions that achieves $G$ from $I$.
\begin{figure*}[ht]
    \centering
    \includegraphics[width=0.8\linewidth]{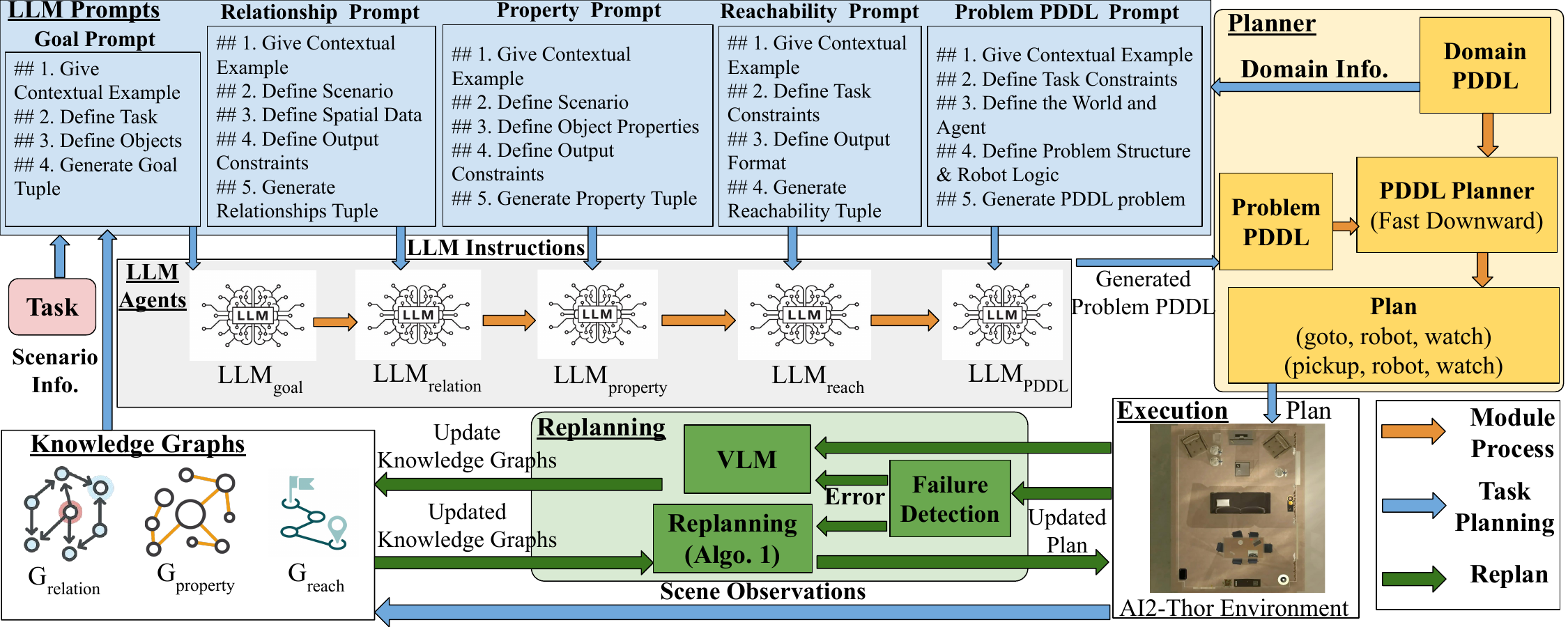}
    \caption{\small Overview of KGLAMP framework. Environment and robot information are encoded as relationship, property, and reachability knowledge graphs. LLM agents generate goal, relational, property, and reachability predicates in a dependency-aware manner to synthesize a PDDL problem, execute the resulting plan, and iteratively update the graphs and replan upon execution failures. }
    \label{fig:Pipeline}
\end{figure*}

\section{Methodology}\label{Sec: Methodology}

In cluttered real-world scenes, even with object inventories, LLM-based PDDL generation from unstructured memory is error-prone, often producing missing constraints, hallucinated predicates, or invalid specifications. Grounding LLMs with a structured knowledge graph organizes entities and relations, reducing context overload and improving reliability. To this end, we introduce \textit{KGLAMP}, a knowledge-graph–guided LLM-based long-horizon planning framework for heterogeneous multi-robot teams. KGLAMP consists of two key components: (1) knowledge graphs that ground the LLM in structured representations of the environment and robot capabilities, and (2) a replanning module that updates plans online when execution deviates from expectations.

As shown in Fig.~\ref{fig:Pipeline}, the framework converts natural language instructions into executable plans using specialized LLM agents. It encodes instructions and visual observations into structured symbolic representations via knowledge graphs. Then, $LLM_{\text{goal}}$, $LLM_{\text{relation}}$, $LLM_{\text{property}}$, and $LLM_{\text{reach}}$ extract goals, spatial relations, object states, and navigation constraints. These are combined to form a PDDL problem, which is solved by a planner (e.g., Fast Downward~\cite{helmert2006fast}) to produce an action sequence. The plan is executed with failure monitoring, and upon failure, a replanning procedure (Alg.~\ref{alg:kg-pddl-correction}) uses a VLM to update the knowledge graph and generate a corrected plan (see Section~\ref{subsec:replanning}).

All LLM modules in our framework ($LLM_{\text{goal}}$, $LLM_{\text{relation}}$, etc.) use the same base model (GPT-5~\cite{openai2025gpt5}) with task-specific prompts. They operate in a sequential pipeline, where each stage produces structured outputs (e.g., goal predicates, relational tuples) that are passed to subsequent modules, ensuring dependency-aware reasoning. Intermediate outputs are validated against the knowledge graph and object states to maintain consistency and improve the correctness of the generated PDDL.

\subsection{Knowledge Graph}
\begin{figure*}[ht]
\centering
\begin{subfigure}[t]{0.28\textwidth}
\centering
\includegraphics[width=1.0\linewidth, height=1.0\linewidth, keepaspectratio]{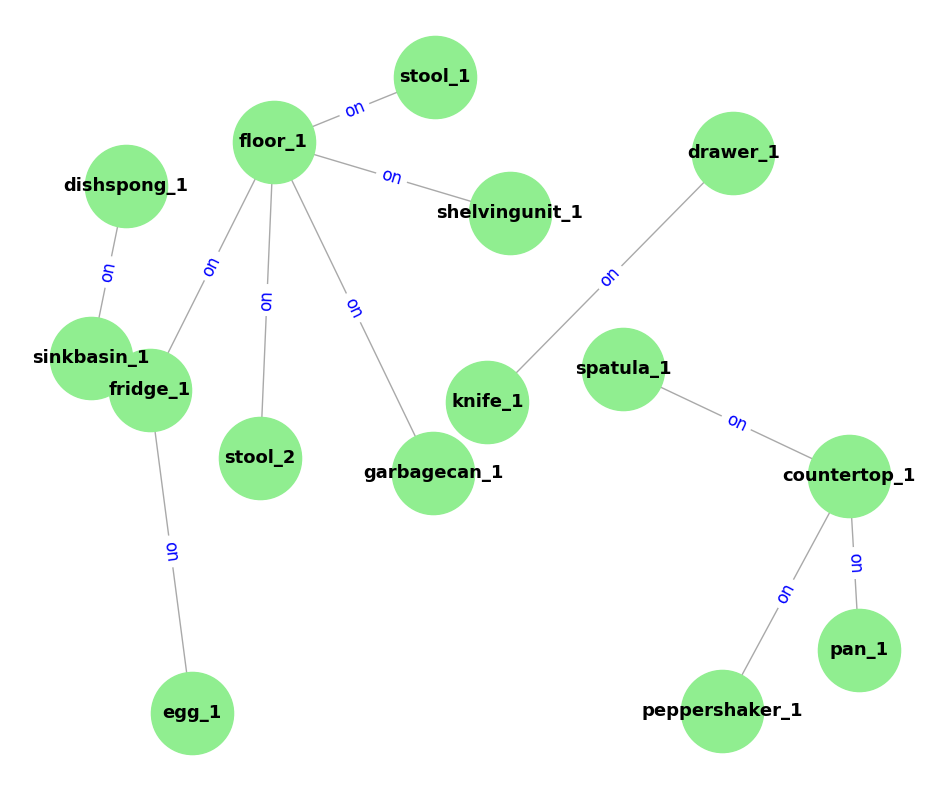}
\caption{Relationship graph $G_{\text{relation}}$}
\label{fig:relational_graph}
\end{subfigure}
~~~
\begin{subfigure}[t]{0.28\textwidth}
\centering
\includegraphics[width=1.0\linewidth, height=1.0\linewidth, keepaspectratio]{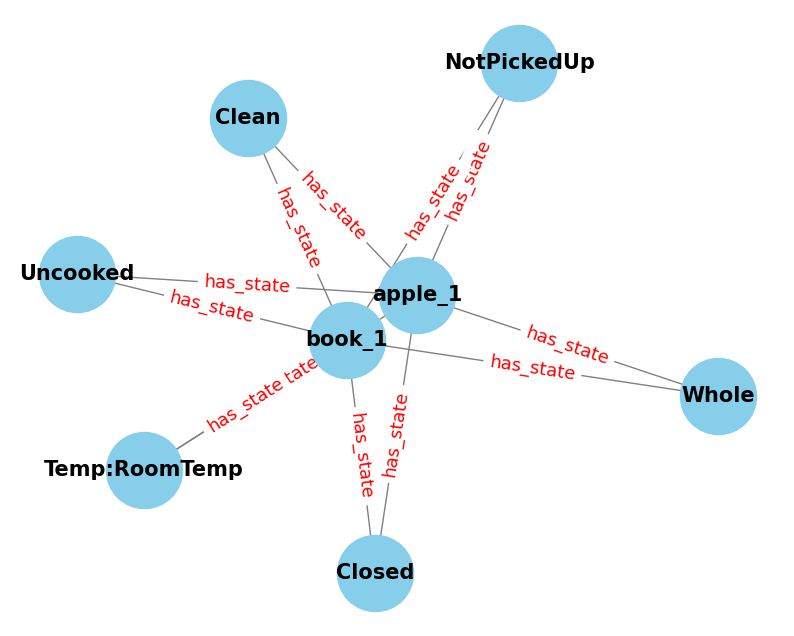}
\caption{Property graph $G_{\text{property}}$}
\label{fig:property_graph}
\end{subfigure}
~~~
\begin{subfigure}[t]{0.28\textwidth}
\centering
\includegraphics[width=1.0\linewidth, height=1.0\linewidth, keepaspectratio]{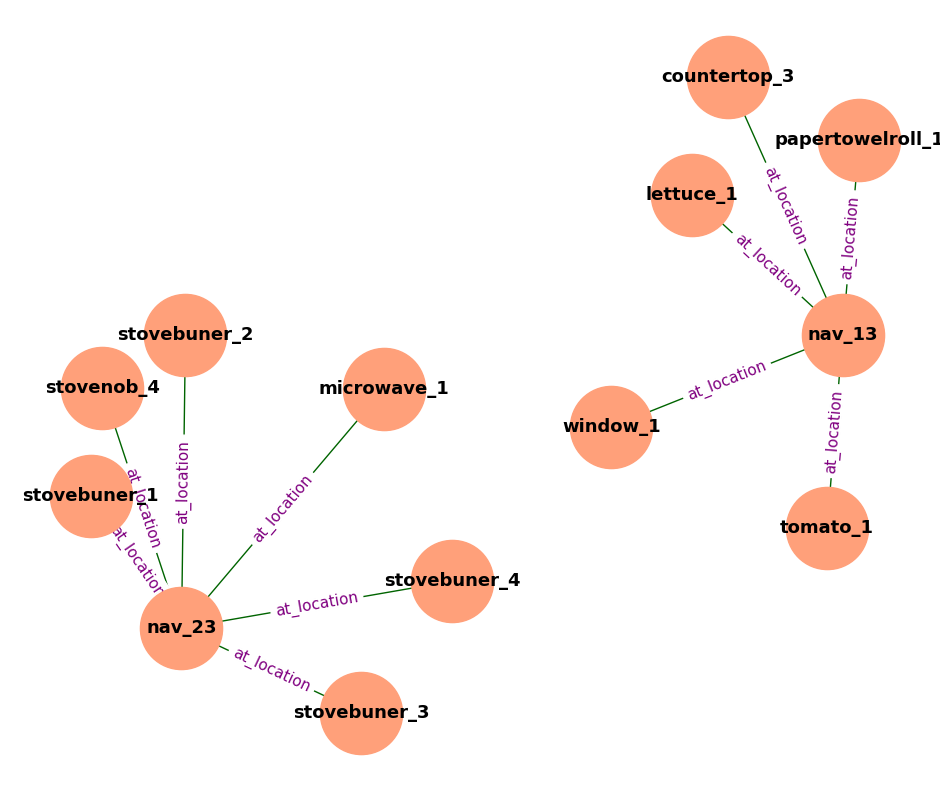}
\caption{Reachability graph $G_{\text{reach}}$}
\label{fig:reachability_graph}
\end{subfigure}

\caption{\small An example knowledge graph. (a) $G_{\text{relation}}$ captures semantic and geometric relationships among objects. (b) $G_{\text{property}}$ encodes object attributes and robot capabilities. (c) $G_{\text{reach}}$  models spatial connectivity. }
\label{fig:all_graphs}
\end{figure*}

A knowledge graph provides a structured representation of world information by encoding entities and their relations. Formally, a knowledge graph $\mathcal{G} = (\mathcal{V}, \mathcal{R}, \mathcal{E})$ consists of entities $\mathcal{V}$, relation types $\mathcal{R}$, and directed edges $\mathcal{E} \subseteq \mathcal{V} \times \mathcal{R} \times \mathcal{V}$ represented as triplets. Each triplet is of the form $(\text{subject}, \text{relation}, \text{object})$ or $(\text{entity}, \text{relation}, \text{property})$, capturing both inter-entity relations and attribute-level information in a unified representation. 

In KGLAMP, the knowledge graph (Fig.~\ref{fig:all_graphs}) is defined as
\begin{equation}
    \mathcal{G} = 
    \mathcal{G}_{\text{relation}} \;\cup\;
    \mathcal{G}_{\text{property}} \;\cup\;
    \mathcal{G}_{\text{reach}},
\end{equation}

\noindent where each component captures a distinct type of relational knowledge for planning. This decomposition simplifies predicate extraction, as task descriptions often omit semantic, physical, or spatial details, and LLMs struggle to infer them jointly. Splitting the graph turns this into smaller, tractable subproblems. Specifically, the Relationship graph, $\mathcal{G}_{\text{relation}}$, captures semantic and geometric object relationships that inform action preconditions and task structure (e.g., $(\texttt{cup}, \texttt{on}, \texttt{table})$); the Property graph, $\mathcal{G}_{\text{property}}$, encodes object attributes and robot capabilities, constraining feasible actions and ensuring that planning respects physical and functional limitations (e.g., $(\texttt{robot\_A},\, $$\texttt{has\_capability},$$\, \texttt{pickup})$); and the Reachability graph, $\mathcal{G}_{\text{reach}}$, links entities to discrete locations, enabling navigability reasoning by grounding goals to reachable target locations (e.g., $(\texttt{microwave}, \texttt{at\_location}, \texttt{location\_1})$).

The initial knowledge graph is constructed from AI2-THOR simulator metadata by extracting object-level attributes such as unique identifiers, semantic labels, spatial positions, and containment relationships. These are converted into triplets, e.g., (cup, in, drawer), forming the basis of the graph. Unique object identifiers ensure consistent entity disambiguation across all components.

\subsection{PDDL Generation Using knowledge graphs}
\begin{figure}
    \centering
    \includegraphics[width=0.75\linewidth]{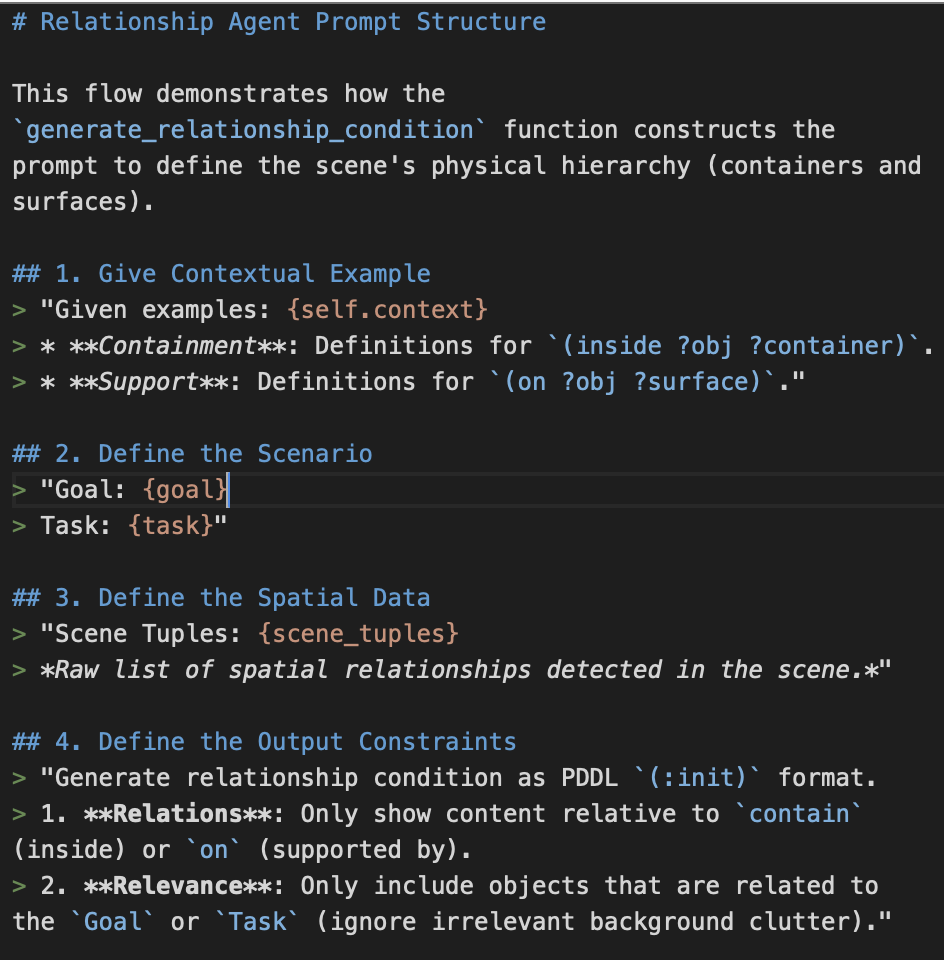}
    \caption{\small An example of LLM prompt for $LLM_\text{relation}$. This prompt utilizes contextual examples, scenario definition, spatial data, and output constraints to extract relevant spatial tuples.}
    \label{fig:prompt}
\end{figure}
Generating correct PDDL is challenging, as small inconsistencies in objects, relations, or properties can invalidate the entire specification. This is amplified in multi-robot settings with ambiguous tasks and complex interactions. To address this, we use a structured pipeline where multiple LLMs incrementally construct the PDDL problem, grounded in the knowledge graph.

\subsubsection{Goal Extraction}
The process begins by translating a natural-language \emph{Task} into a formal goal specification. Since human instructions are often ambiguous or misaligned with the environment, contextual grounding is required. Conditioning $\text{LLM}{\text{goal}}$ on the environment \emph{Objects} produces a goal consistent with the available entities, attributes, and relations:
 \begin{equation}
\textit{Goal} = \text{LLM}_{\text{goal}}(\textit{Task}, \textit{Objects}).
\end{equation}

\subsubsection{Relationship Extraction}
After extracting the goal, $\text{LLM}_{\text{relation}}$ infers object–object relations needed for planning. These relations capture structural dependencies and constrain action feasibility and ordering (e.g., reachability or containment). Identifying these relations before action generation is therefore critical, as they define the structural constraints governing valid action sequences. This inference is grounded in $\mathcal{G}_{\text{relation}}$, yielding: 
\begin{equation}
    \textit{Relation} = \text{LLM}_{\text{relation}}(\textit{Goal},\, \textit{Task},\, \mathcal{G}_{\text{relation}}).
\end{equation}
\subsubsection{Property Assignment}
With relational structure defined, $\text{LLM}_{\text{property}}$ assigns object attributes and robot capabilities that constrain feasible actions. These properties determine action participation and whether robots have the required skills to achieve the goal. Accurate assignment is critical to avoid infeasible actions and ensure goal achievability. This inference step leverages $\mathcal{G}_{\text{property}}$, yielding: 
\begin{equation}
    \textit{Property} = \text{LLM}_{\text{property}}(\textit{Goal},\, \textit{Task},\, \textit{Relation},\, \mathcal{G}_{\text{property}}).
\end{equation}

\subsubsection{Navigation Structure}

We decouple navigation from object interaction, as jointly reasoning over spatial motion and manipulation constraints substantially increases planning complexity. Navigation reasoning concerns spatial reachability, path feasibility, and inter-robot interference, while object interaction involves manipulation-specific preconditions and effects. Jointly inferring these heterogeneous constraints often yields incomplete or inconsistent predicates. By isolating navigation reasoning, $\text{LLM}_{\text{reach}}$ focuses on identifying accessible locations and feasible robot motions while avoiding conflicts. Grounded in $\mathcal{G}_{\text{reach}}$, which encodes spatial layout and connectivity, this step produces navigation predicates with greater consistency and reliability:
\begin{equation}
    \textit{Reach} = \text{LLM}_{\text{reach}}(\textit{Relation},\, \textit{Property},\, \mathcal{G}_{\text{reach}}).
\end{equation}

\subsubsection{Problem PDDL}
Finally, the goal, relational structure, object properties, and navigation constraints are integrated to construct the PDDL problem. $\text{LLM}_{\text{PDDL}}$ synthesizes a coherent symbolic representation that captures both environment constraints and robot capabilities:
\begin{equation}
    \text{Prob}_{\text{PDDL}} = \text{LLM}_{\text{PDDL}}(\textit{Goal},\, \textit{Relation},\, \textit{Property},\, \textit{Reach}).
\end{equation}
This structured workflow ensures each PDDL component is explicitly grounded rather than implicitly inferred. By decomposing goal, relation, property, and navigation extraction, it reduces common inconsistencies such as missing predicates, invalid preconditions, and incorrect object bindings. As a result, the generated PDDL more accurately reflects the environment and agent capabilities, producing executable and coherent plans, as evidenced by the results in Section~\ref{Section: Experiments}.

\subsubsection{Prompt Design}
Prompt design follows five key principles (see Fig.~\ref{fig:prompt}). First, clear task definition specifies the objective explicitly. Second, structured output formats enforce consistency in model responses. Third, explicit constraints bound feasible reasoning and define logical and spatial limits. Fourth, step-by-step decomposition guides multi-stage inference for complex tasks. Fifth, few-shot examples improve robustness and reduce ambiguity.

\subsection{Replanning for Symbolic Execution Failures}\label{subsec:replanning}
Symbolic planners are highly sensitive to inconsistencies in domain specifications, where even minor errors in PDDL can cause failure. Such issues often stem from missing preconditions, affordances, or state transitions. For instance, encoding a box as \texttt{(box, has\_state, closed)} without specifying that it is openable prevents the planner from retrieving its contents. These failures are common in long-horizon tasks, motivating the need for a robust replanning mechanism to detect and correct incomplete or inconsistent representations.

To address these failures, we first diagnose execution errors by extracting an error signal $e$, indicating planner failure (no plan), execution failure (violated preconditions), or perception failure (missing objects). Conditioned on $e$, we invoke the refinement procedure (Alg.~\ref{alg:kg-pddl-correction}), where the LLM proposes candidate updates to the knowledge graph:
\begin{equation}
\{H_1, \ldots, H_k\}, \;
\{\pi_1, \ldots, \pi_k\}
\;=\;
\mathrm{LLM}_{\text{replan}}(k, e, \mathcal{G}).
\end{equation}
Each $H_i$ represents a potential correction (e.g., adding missing affordances, states, or relations), $k$ is the number of candidates, and $\pi_i$ denotes its estimated likelihood of that candidate.

Each hypothesis is evaluated by updating the knowledge graph, regenerating the PDDL domain and problem, and re-running the planner. If a valid plan is found, we compute the plan cost difference $\Delta c_i$; otherwise, the candidate is discarded.
Selection is performed using the probability–cost objective: 
\begin{equation}
H^\star \;=\;
\arg\max_i \frac{p_i}{(\Delta c_i)^{\lambda}},
\end{equation}
where $p_i = \pi_i / \sum_{j=1}^k \pi_j$ is the normalized candidate probability. The parameter $\lambda$ balances plausibility and minimal symbolic change (we use $\lambda=2$). Plan cost is defined as PDDL plan length, providing a simple, domain-agnostic measure of complexity without bias from parallel execution. Replanning stops when a valid plan is found or after three iterations. For partial observability, a CLIP-based VLM processes egocentric RGB observations.

By coupling failure diagnosis, LLM-guided repair, and planner-in-the-loop validation, the replanning module enables recovery from symbolic inconsistencies and improves reliability in long-horizon manipulation tasks.


\subsection{VLM-Based Discovery under Partial Observability}
Due to partial observability, the initial knowledge graph may be incomplete. To address this, we use a VLM to analyze visual observations after execution or upon failure. The VLM detects previously unseen entities and infers their spatial relations, which are converted into semantic tuples and added to the knowledge graph. These updates enable dynamic replanning, allowing the symbolic planner to incorporate newly discovered information and recover from incomplete world states.

\begin{algorithm}[t]
\centering
\begin{algorithmic}[1]
    
    \STATE Initialize iteration counter $t \gets 0,$ error $ e,$ number of candidates $k,$ knowledge graph $ \mathcal{G},$ trade-off  coefficient $ \lambda$, domain $d$, problem PDDL $\text{Prob}_{PDDL}$ 
    \WHILE{$e \neq \emptyset$}
        \STATE $\{H_1,\ldots,H_k\}, \{\pi_1,\ldots,\pi_k\}
               \gets \mathrm{LLM}_{\text{replan}}(k,e,\mathcal{G})$ \blue{ \COMMENT{top-$k$ candidate fixes using current error}}

        \STATE \textbf{for} $i = 1,\ldots,k$ \textbf{do}
        \STATE \hspace{1.2em} $p_i \gets \pi_i \big/ \displaystyle\sum_{j=1}^k \pi_j$  \blue{\COMMENT{normalize probabilities}}
        \STATE \textbf{end for}
        \FOR{$i = 1$ \TO $k$}
            \STATE $\mathcal{G}^{(i)}_{\mathrm{temp}} \gets \mathcal{G} \cup H_i$ \blue{ \COMMENT{update knowledge graph}}
            \STATE $\text{Prob}_{PDDL, i}^{\mathrm{new}} \gets
                   \textsc{GeneratePDDL}(\mathcal{G}^{(i)}_{\mathrm{temp}})$ \blue{\COMMENT{get PDDL}}
            \STATE $\textit{plan}_i, e_i \gets
                   \textsc{Planner}(d, \text{Prob}_{PDDL, i}^{\mathrm{new}})$
            \blue{\COMMENT{get new plan; capture new error message if any}}
            \IF{$\textit{plan}_{i}$ is valid} 
                \STATE $\Delta c_i \gets
                       \mathrm{cost}(d, \text{Prob}_{PDDL, i}^{\mathrm{new}})
                       - \mathrm{cost}(d,\text{Prob}_{PDDL})$ \blue{ \COMMENT{change in cost evaluation}}
            \ELSE
                \STATE $\Delta c_i \gets \infty$
            \ENDIF
        \ENDFOR

        \STATE $H^\star \gets
               \displaystyle \arg\max_i \frac{p_i}{\left(\Delta c_i\right)^{\lambda}}$
        \blue{\COMMENT{best candidate selection}}

        \STATE $\mathcal{G} \gets \mathcal{G} \cup H^\star$ \blue{ \COMMENT{updated knowledge graph}}
        \STATE $\text{Prob}_{PDDL} \gets \textsc{Generate PDDL}(\mathcal{G})$ \blue{\COMMENT{get final PDDL}}
        \STATE $\textit{plan}, e \gets \textsc{Planner}(d,\text{Prob}_{PDDL})$ \blue{\COMMENT{get updated plan}}
    \ENDWHILE
    \RETURN $\mathcal{G}, \text{Prob}_{PDDL}, \textit{plan}$
\end{algorithmic}
    \caption{\small KGLAMP Replanning} 
   \label{alg:kg-pddl-correction}
\end{algorithm}

\section{Experiments}\label{Section: Experiments}
In this section, we provide details of our experiments.

\subsection{Dataset}
We evaluate our framework on the MAT2-THOR benchmark~\cite{gupta2026scale}, a multi-agent long-horizon dataset built on AI2-THOR~\cite{kolve2017ai2}. It includes $49$ indoor multi-robot tasks across seven floor plans, categorized as: (i) $25$ \emph{Simple Tasks} with short horizons and one to two subgoals (e.g., ``Open the laptop and turn it on”); (ii) $17$ \emph{Complex Tasks} requiring coordinated execution by heterogeneous robots (e.g., ``Slice the lettuce, trash the mug, and switch off the light”); and (iii) $7$ \emph{Vague Command Tasks} with underspecified instructions (e.g., ``Prepare ingredients for cooking a sandwich tomorrow”).

\subsection{Qualitative Analysis: Symbolic Replanning}
\begin{figure}[t]
    \centering
    \includegraphics[width=1\linewidth, height=1\linewidth, keepaspectratio]{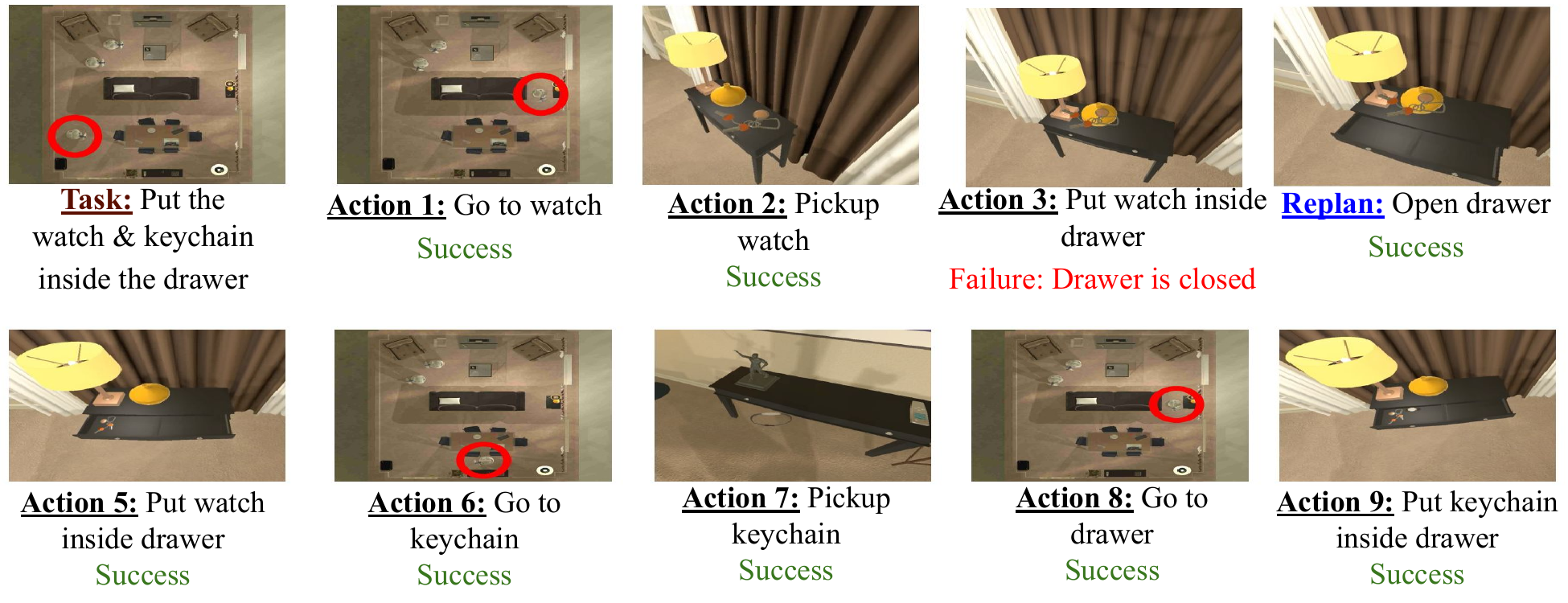}
    \caption{\small Qualitative example of planning and replanning. In the task \textit{Put the watch and keychain inside the drawer}, the robot fails when placing the watch into a closed drawer. It recovers by replanning to open the drawer and completes the task.}
    \label{fig:Quality}
\end{figure}

An example of replanning improving robustness is shown in Fig.~\ref{fig:Quality} for the task: \textit{Put the watch and keychain inside the drawer}. The goal is encoded as $(\text{watch}, \text{in}, \text{drawer})$ and $(\text{keychain}, \text{in}, \text{drawer})$. Execution fails because the model omits the precondition that the drawer must be open, causing failure at \textit{Action 3}. The system identifies the missing precondition, updates the knowledge graph, and inserts a drawer-opening action. The repaired plan then succeeds, illustrating effective recovery through symbolic replanning.

\subsection{Evaluation Metrics and Baselines}
We utilize five evaluation metrics that capture task success, action validity, and temporal efficiency. \emph{Task Completion Rate (TCR)} measures the percentage of tasks in which all goal conditions are satisfied. \emph{Goal Condition Recall (GCR)} quantifies the proportion of ground-truth goal conditions achieved in the final state. \emph{Executability Rate (ER)} reports the percentage of planned actions that are successfully executed in the simulator, independent of task relevance. \emph{Planning Time (PT)} denotes the time required to generate the final multi-robot plan, while \emph{Execution Time (ET)} measures the time to execute the plan in AI2-THOR. Together, these metrics are widely used in the literature~\cite{zhang2025lamma, kannan2024smart} and assess correctness, robustness, and operational efficiency.

We compare against six representative baselines. (i) \emph{LLM-as-Planner}~\cite{zhou2023large} treats the language model as a standalone planner that directly generates action sequences without structural constraints or verification. (ii) \emph{LLM-as-Planner with Chain-of-Thought (CoT)}~\cite{wei2022chain} elicits explicit reasoning before plan generation, with plans manually converted into executable actions. (iii) \emph{LLM+P}~\cite{liu2023llm+} adds a post-hoc execution-phase validator that detects invalid actions but provides no planning-time guidance. (iv) \emph{LaMMA-P}~\cite{zhang2025lamma} incorporates structured knowledge and adaptive multi-agent reasoning to improve plan coherence and execution reliability. (v) \emph{SayPlan}~\cite{rana2023sayplan} leverages structured scene graphs to ground LLM reasoning, but relies on static graph representations. (vi) \emph{HVR}~\cite{hvr_icml_2025} employs retrieval-augmented generation (RAG) to incorporate relevant textual knowledge, but depends on unstructured retrieved context.  We set a 300-second planning time limit, and use GPT-5~\cite{openai2025gpt5} for all methods to ensure a fair comparison.

\subsection{Performance in Full Knowledge Graph}

\begin{table*}[t]
\centering
\resizebox{\textwidth}{!}{
\begin{tabular}{l|ccccc|ccccc|ccccc}
\hline
\textbf{Method} 
& \multicolumn{5}{c|}{\textbf{Basic}} 
& \multicolumn{5}{c|}{\textbf{Complex}} 
& \multicolumn{5}{c}{\textbf{Vague}} \\
\cline{2-16}
& \textbf{TCR(\%)} & \textbf{GCR(\%)} & \textbf{ER(\%)} & \textbf{PT(s)} & \textbf{ET(s)}
& \textbf{TCR(\%)} & \textbf{GCR(\%)} & \textbf{ER(\%)} & \textbf{PT(s)} & \textbf{ET(s)}
& \textbf{TCR(\%)} & \textbf{GCR(\%)} & \textbf{ER(\%)} & \textbf{PT(s)} & \textbf{ET(s)} \\
\hline
LLM-as-Planner               
& 32.0 & 44.0 & 76.0 & 32.6 & 45.0 
& 10.5 & 22.8 & 69.8 & 39.9 & 55.9 
& 14.3 & 21.4 & 12.1 & 53.5 & 106.6 \\

LLM-as-Planner (CoT)         
& 36.0 & 44.0 & 77.2 & \textbf{27.6} & 54.0 
& 26.3 & 46.5 & 77.9 & \textbf{36.9} & 69.2 
& 14.3 & 21.4 & 35.7 & 43.1 & 35.2 \\

LLM + P                
& 20.0 & 36.0 & 60.1 & 61.1 & \textbf{35.4} 
& 15.8 & 32.4 & 15.9 & 83.1 & \textbf{52.6} 
& 0.0 & 0.0 & 0.0 & \textbf{10.7} & \textbf{0.0} \\

LaMMA-P                
& 60.0 & 74.0 & 90.0 & 131.8 & 114.8 
& 36.8 & 60.1 & 74.3 & 190.9 & 163.9 
& 57.1 & 57.1 & 63.7 & 167.7 & 199.0 \\

SayPlan 
& 65.0 & 74.0 & 92.0 & 4.7 & 70.9
& 27.0 & 55.0 & 83.0 & 5.6 & 79.5
& 29.0 & 29.0 & 80.0 & 5.3 & 50.1 \\

HVR
& 60.0 & 76.0 & 88.0 & 54.1 & 84.1
& 24.0 & 55.0 & 81.0 & 73.8 & 95.2
& 43.0 & 60.0 & 89.0 & 94.4 & 88.1 \\

Ours w/o replan     
& 72.0 & 80.0 & 91.3 & 171.4 & 95.5 
& 47.4 & 52.6 & 68.9 & 236.6 & 89.4 
& 57.1 & 57.1 & 92.9 & 169.1 & 172.2 \\

Ours   
& \textbf{84.0} & \textbf{85.2} & \textbf{97.5} & 181.7 & 67.7 
& \textbf{68.4} & \textbf{60.3} & \textbf{94.9} & 284.3 & 144.8 
& \textbf{71.4} & \textbf{71.4} & \textbf{94.3} & 301.5 & 77.4 \\
\hline
\end{tabular}}
\caption{\small Performance comparison across methods.}
\label{tab:exp1}
\end{table*}
As shown in Table~\ref{tab:exp1}, our method achieves the highest task completion rate, outperforming the strongest baseline, \emph{LaMMA-P}, by $25.3\%$ across all $49$ tasks.
We compare against four categories of methods. The first includes \emph{LLM-as-Planner} and \emph{LLM-as-Planner + CoT}, which directly generate action sequences from task descriptions. While efficient, they struggle with complex or long-horizon tasks due to error accumulation and lack of grounding in environmental structure and constraints.

The second category, \emph{LLM+P}, uses an LLM to generate a PDDL problem and a classical planner for execution. While effective when the PDDL is accurate, it often fails due to invalid or incomplete specifications, such as incorrect predicates, action schemas, or object definitions, leading to planner failure or infeasible plans.

The third category, \emph{LaMMA-P}, introduces explicit task decomposition, translating each subtask into a separate PDDL problem. Although this improves scalability, it does not explicitly model inter-agent coordination or object-level dependencies across subtasks. Consequently, when task preconditions depend on other agents’ actions or shared object states, the resulting plans can be inconsistent or incomplete.

The fourth category integrates graph-based grounding and retrieval-augmented reasoning, exemplified by \emph{SayPlan} and \emph{HVR}. While these methods improve grounding on simpler tasks, their reliance on static graphs or unstructured retrieval limits multi-step reasoning and adaptability, leading to degraded performance in complex, long-horizon scenarios.

In contrast, our method uses a unified framework that jointly coordinates all robots while explicitly modeling object-level dependencies. By leveraging a knowledge graph capturing semantic, spatial, and relational structure, it enables grounded reasoning and reliable execution, achieving the highest overall success rate.

\subsection{Heterogeneity Evaluation}

To evaluate heterogeneity-aware planning, we focus on tasks requiring coordination among robots with distinct capabilities. In MAT2-THOR, $19$ out of $49$ tasks involve heterogeneous robots with disjoint skill sets, making capability-aware task allocation necessary for successful execution.

In our framework, heterogeneity is encoded via the property graph $G_{\text{property}}$, which represents robot-specific capabilities and constrains action feasibility during PDDL generation, enabling implicit capability-conditioned task allocation.

On these heterogeneous tasks, our method achieves a TCR of $73.7\%$, consistent with overall performance, demonstrating robustness to capability asymmetry. In contrast, baselines that assume shared action models often produce infeasible plans by assigning actions to incapable robots.

Qualitatively, in the task ``slice the lettuce and put it in the fridge," we observe three robots with  limited skills that make their roles unique and irreplaceable.  The slicing action is strictly assigned to the specific robot possessing that capability, transport is handled by the other robot, and the third is left unassigned, illustrating clear role specialization where agents cannot substitute for one another.

\subsection{Ablation Study}
\begin{table}[t]
\centering
\resizebox{0.48\textwidth}{!}{\begin{tabular}{lccccc}
\hline
\textbf{Method} & \textbf{TCR(\%)} & \textbf{GCR(\%)} & \textbf{ER(\%)} & \textbf{PT(s)} & \textbf{ET(s)} \\
\hline
Ours             
& \textbf{76.5} & \textbf{74.0} & \textbf{96.1} & 236.4 & 97.75 \\

Ours w/o replan
& 60.7 & 60.2 & 83.1 & 195.4 & 103.7 \\

Ours w/o $\mathcal{G}_{\text{reach}}$   
& 58.8 & 64.7 & 83.8 & 181.7 & 99.0 \\

Ours w/o $\mathcal{G}_{\text{reach}}$ and $\mathcal{G}_{\text{property}}$
& 52.9 & 60.2 & 80.6 & 156.5 & 83.7 \\

Ours w/o $\mathcal{G}_{\text{reach}}$, $\mathcal{G}_{\text{property}}$ and $\mathcal{G}_{\text{relation}}$
& 37.2 & 48.4 & 68.6 & \textbf{95.6} & \textbf{76.8} \\

\hline
\end{tabular}}
\caption{\small Ablation results with different graph structures.
}
\label{tab:exp2}
\end{table}
As shown in Table~\ref{tab:exp2}, the ablation study shows that each component is critical and non-redundant. Removing any module degrades performance, with the largest drop occurring when all graph structures are removed, highlighting the importance of structured semantic, property, and relational information for guiding manipulation.

Specifically, $\mathcal{G}_{\text{property}}$ identifies which objects must be manipulated, $\mathcal{G}_{\text{relation}}$ encodes affordances for action selection, and $\mathcal{G}_{\text{reach}}$ enforces spatial and relational consistency to support correct coordinate-frame reasoning and collision avoidance. The replanning module further improves robustness by enabling recovery from execution failures and partial observability, such as when a target object is initially enclosed and must first be exposed. Across the evaluation set, replanning is triggered in $12$ out of $49$ tasks ($\sim 24.5\%$), indicating that execution failures and symbolic inconsistencies are common in long-horizon settings. This highlights the practical importance of the replanning module, which plays an active role in recovering from errors and ensuring successful task completion. This behavior is particularly critical in complex tasks (Table I), where failure recovery significantly contributes to improved task completion rates. 

\subsection{Symbolic Consistency Analysis}
To assess symbolic quality, we measure planner success rate, defined as the percentage of generated PDDL problems solvable by the planner. Our method achieves $86.2\%$, indicating that most PDDL specifications are syntactically and semantically valid. Remaining failures stem from incomplete or inconsistent representations, such as missing critical objects or incorrect preconditions. This analysis highlights the effectiveness of knowledge-graph grounding in improving symbolic consistency, while also identifying areas for further improvement in handling ambiguous or partially observed environments.

\subsection{Task Planning under Partial Observability}
To evaluate robustness under environmental uncertainty, we consider partial observability by omitting 12 critical objects (e.g., \textit{Tomato}, \textit{Laptop}) from the initial world model, requiring VLM-driven object discovery and replanning. Our method achieves a TCR of $64.0\%$ and a GCR of $74.0\%$, substantially outperforming \emph{LaMMA-P}, which attains a TCR of $12.0\%$ and a GCR of $36.1\%$ due to plan inconsistency.

These results demonstrate the effectiveness of our replanning mechanism in dynamically updating the knowledge graph as new entities are perceived. The remaining gap relative to full-observability settings highlights challenges in active perception, where VLM recognition errors and suboptimal viewpoints can hinder reliable object identification in cluttered scenes. Despite these limitations, integrating VLM feedback into graph-based planning yields significantly more robust long-horizon task execution than existing baselines.


\subsection{Local Environment Model Comparison}
We evaluate several open-source language models in a controlled local setting using the \texttt{ollama} framework. All experiments are conducted on an Ubuntu~22.04 workstation with an AMD~Ryzen~Threadripper~7960X CPU, 128~GB RAM, and an NVIDIA~GeForce~RTX~5080 GPU. The evaluated models include Llama~3.2~3B~\cite{llama3herd2024}, Phi~3~Mini~7B~\cite{abdin2024phi}, Mistral~7B~\cite{jiang2023mistral7b}, and Qwen~2~7B~\cite{yang2024qwen2}.
\begin{table}[t]
\centering
\begin{tabular}{lccccc}
\hline
\textbf{Method} & \textbf{TCR(\%)} & \textbf{GCR(\%)} & \textbf{ER(\%)} & \textbf{PT(s)} & \textbf{ET(s)} \\
\hline
llama3.2 3B               
& 39.2 & 53.3 & 71.0 & 24.9 & 67.5 \\

phi3:mini 7B   
& 41.2 & 53.3 & 72.9 & \textbf{13.9} & 86.4 \\

mistral 7B
& 41.2 & 54.9 & 73.0 & 16.5 & \textbf{59.6} \\

qwen2:7B
& \textbf{45.0} & \textbf{56.9} & \textbf{73.2} & 18.8 & 80.5 \\
\hline
\end{tabular}
\caption{\small Performance comparison across models.} 
\label{tab:exp3}
\end{table}
As shown in Table~\ref{tab:exp3}, the models exhibit broadly similar performance across tasks. Qwen~2~7B achieves the strongest results, with a $5.8\%$ higher TCR than Llama~3.2, though the gap remains modest, indicating limited sensitivity to model choice under identical hardware and runtime conditions.


Inference efficiency is comparable across models, with all achieving response times under $25$ seconds, indicating practical responsiveness for local deployment. Their smaller sizes (3B–7B) relative to frontier models like GPT-5 (\textgreater100B parameters) contribute to this efficiency. However, as these models are not optimized for complex reasoning, their performance may be limited on more challenging tasks.




\section{Conclusion}\label{Section: Conclusions}

We presented \textit{KGLAMP}, a knowledge-graph–guided LLM planning framework for long-horizon planning in heterogeneous multi-robot systems. By grounding planning in unified structured graphs encoding object relations, spatial reachability, and robot capabilities, \textit{KGLAMP} enables LLMs to generate task representations that better reflect real-world constraints. Experiments show that \textit{KGLAMP} substantially outperforms LLM-only and PDDL-based baselines, achieving a $25.3\%$ improvement in task completion rate and a $15.3\%$ reduction in execution failures relative to the strongest baseline across all tasks. Moreover, incremental knowledge-graph updates improve robustness by enabling adaptation to incomplete or evolving environment information, maintaining performance in dynamic settings.

Overall, \textit{KGLAMP} demonstrates that coupling LLM-based reasoning with structured, continuously updated knowledge representations is a promising direction for reliable, scalable, and context-aware planning in heterogeneous multi-robot teams. Future work will explore deployment on physical robots, training smaller distilled models via knowledge distillation~\cite{gupta2024towards}, tighter integration with perception and feedback loops, and scaling to increasingly unstructured environments.

\section{Limitations}\label{Section: Limitations}
Despite its effectiveness, our approach has several limitations. First, it relies on sufficiently detailed environmental and task information to construct the knowledge graph; while LLMs can infer missing attributes, such inference may introduce noise or inconsistencies. Second, coordinating multiple LLM components incurs significant computational overhead, limiting scalability for real-time or large-scale deployment. Third, even with explicitly specified domain symbols and type definitions, general-purpose LLMs may generate PDDL predicates or action schemas that violate syntactic or semantic constraints, yielding invalid planning domains without additional verification or correction mechanisms.

\bibliographystyle{unsrt}
\bibliography{refs}

@INPROCEEDINGS{11128711,
  author={Gupta, Piyush and Isele, David and Sachdeva, Enna and Huang, Pin-Hao and Dariush, Behzad and Lee, Kwonjoon and Bae, Sangjae},
  booktitle={2025 IEEE International Conference on Robotics and Automation (ICRA)}, 
  title={Generalized Mission Planning for Heterogeneous Multi-Robot Teams via {LLM}-Constructed Hierarchical Trees}, 
  year={2025},
  volume={},
  number={},
  pages={10187-10193},
  doi={10.1109/ICRA55743.2025.11128711}}

@inproceedings{kang2025gflowvlm,
  title={{GFlowVLM}: Enhancing Multi-step Reasoning in Vision-Language Models with Generative Flow Networks},
  author={Kang, Haoqiang and Sachdeva, Enna and Gupta, Piyush and Bae, Sangjae and Lee, Kwonjoon},
  booktitle={Proceedings of the Computer Vision and Pattern Recognition Conference},
  pages={3815--3825},
  year={2025}
}

@article{gupta2025graph,
  title={Graph-Grounded {LLM}s: Leveraging Graphical Function Calling to Minimize {LLM} Hallucinations},
  author={Gupta, Piyush and Bae, Sangjae and Isele, David},
  journal={arXiv preprint arXiv:2503.10941},
  year={2025}
}

@inproceedings{zhang2024multi,
  title={Multi-robot coordination and layout design for automated warehousing},
  author={Zhang, Yulun and Fontaine, Matthew C and Bhatt, Varun and Nikolaidis, Stefanos and Li, Jiaoyang},
  booktitle={Proceedings of the International Symposium on Combinatorial Search},
  volume={17},
  pages={305--306},
  year={2024}
}

@inproceedings{gupta2024towards,
  title={Towards scalable \& efficient interaction-aware planning in autonomous vehicles using knowledge distillation},
  author={Gupta, Piyush and Isele, David and Bae, Sangjae},
  booktitle={2024 IEEE Intelligent Vehicles Symposium (IV)},
  pages={2735--2742},
  year={2024},
  organization={IEEE}
}

@inproceedings{bai2021multi,
  title={Multi-robot task planning under individual and collaborative temporal logic specifications},
  author={Bai, Ruofei and Zheng, Ronghao and Liu, Meiqin and Zhang, Senlin},
  booktitle={International Conference on Intelligent Robots and Systems (IROS)},
  pages={6382--6389},
  year={2021},
  organization={IEEE}
}

@article{liu2024autonomous,
  title={Autonomous Robot Task Execution in Flexible Manufacturing: Integrating {PDDL} and Behavior Trees in {ARIAC} 2023},
  author={Liu, Ruikai and Wan, Guangxi and Jiang, Maowei and Chen, Haojie and Zeng, Peng},
  journal={Biomimetics},
  volume={9},
  number={10},
  pages={612},
  year={2024},
  publisher={MDPI}
}

@article{helmert2006fast,
  title={The fast downward planning system},
  author={Helmert, Malte},
  journal={Journal of Artificial Intelligence Research},
  volume={26},
  pages={191--246},
  year={2006}
}

@inproceedings{aineto2018learning,
  title={Learning {STRIPS} action models with classical planning},
  author={Aineto, Diego and Jim{\'e}nez, Sergio and Onaindia, Eva},
  booktitle={Proceedings of the International Conference on Automated Planning and Scheduling},
  volume={28},
  pages={399--407},
  year={2018}
}

@article{gupta2022incentivizing,
  title={Incentivizing collaboration in heterogeneous teams via common-pool resource games},
  author={Gupta, Piyush and Bopardikar, Shaunak D and Srivastava, Vaibhav},
  journal={IEEE Transactions on Automatic Control},
  volume={68},
  number={3},
  pages={1902--1909},
  year={2022},
  publisher={IEEE}
}

@inproceedings{gupta2019achieving,
  title={Achieving efficient collaboration in decentralized heterogeneous teams using common-pool resource games},
  author={Gupta, Piyush and Bopardikar, Shaunak D and Srivastava, Vaibhav},
  booktitle={58th Conference on Decision and Control (CDC)},
  pages={6924--6929},
  year={2019},
  organization={IEEE}
}

@article{ren2023robots,
  title={Robots that ask for help: Uncertainty alignment for large language model planners},
  author={Ren, Allen Z and Dixit, Anushri and Bodrova, Alexandra and Singh, Sumeet and Tu, Stephen and Brown, Noah and Xu, Peng and Takayama, Leila and Xia, Fei and Varley, Jake and others},
  journal={arXiv preprint arXiv:2307.01928},
  year={2023}
}

@article{liu2023llm+,
  title={{LLM+P}: Empowering large language models with optimal planning proficiency},
  author={Liu, Bo and Jiang, Yuqian and Zhang, Xiaohan and Liu, Qiang and Zhang, Shiqi and Biswas, Joydeep and Stone, Peter},
  journal={arXiv preprint arXiv:2304.11477},
  year={2023}
}

@inproceedings{zhang2025lamma,
  title={{LaMMA-P}: Generalizable multi-agent long-horizon task allocation and planning with {LM}-driven {PDDL} planner},
  author={Zhang, Xiaopan and Qin, Hao and Wang, Fuquan and Dong, Yue and Li, Jiachen},
  booktitle={International Conference on Robotics and Automation},
  pages={10221--10221},
  year={2025},
  organization={IEEE}
}

@article{gong2025zero,
  title={Zero-Shot Iterative Formalization and Planning in Partially Observable Environments},
  author={Gong, Liancheng and Zhu, Wang and Thomason, Jesse and Zhang, Li},
  journal={arXiv preprint arXiv:2505.13126},
  year={2025}
}

@article{kolve2017ai2,
  title={{AI2-THOR}: An interactive {3D} environment for visual {AI}},
  author={Kolve, Eric and Mottaghi, Roozbeh and Han, Winson and VanderBilt, Eli and Weihs, Luca and Herrasti, Alvaro and Deitke, Matt and Ehsani, Kiana and Gordon, Daniel and Zhu, Yuke and others},
  journal={arXiv preprint arXiv:1712.05474},
  year={2017}
}

@article{zhou2023large,
  title={Large language model as a policy teacher for training reinforcement learning agents},
  author={Zhou, Zihao and Hu, Bin and Zhao, Chenyang and Zhang, Pu and Liu, Bin},
  journal={arXiv preprint arXiv:2311.13373},
  year={2023}
}

@article{wei2022chain,
  title={Chain-of-thought prompting elicits reasoning in large language models},
  author={Wei, Jason and Wang, Xuezhi and Schuurmans, Dale and Bosma, Maarten and Xia, Fei and Chi, Ed and Le, Quoc V and Zhou, Denny and others},
  journal={Advances in neural information processing systems},
  volume={35},
  pages={24824--24837},
  year={2022}
}

@article{goel2024novelgym,
  title={{NOVELGYM}: A Flexible Ecosystem for Hybrid Planning and Learning Agents Designed for Open Worlds},
  author={Goel, Shivam and Wei, Yichen and Lymperopoulos, Panagiotis and Chur{\'a}, Kl{\'a}ra and Scheutz, Matthias and Sinapov, Jivko},
  journal={arXiv preprint arXiv:2401.03546},
  year={2024}
}

@article{ferreira2024distributed,
  title={Distributed allocation and scheduling of tasks with cross-schedule dependencies for heterogeneous multi-robot teams},
  author={Ferreira, Barbara Arbanas and Petrovi{\'c}, Tamara and Orsag, Matko and Mart{\'\i}nez-de Dios, J Ramiro and Bogdan, Stjepan},
  journal={IEEE access},
  volume={12},
  pages={74327--74342},
  year={2024},
  publisher={IEEE}
}

@article{cao2023robot,
  title={Robot behavior-tree-based task generation with large language models},
  author={Cao, Yue and Lee, CS},
  journal={arXiv preprint arXiv:2302.12927},
  year={2023}
}

@article{yuan2023skill,
  title={Skill reinforcement learning and planning for open-world long-horizon tasks},
  author={Yuan, Haoqi and Zhang, Chi and Wang, Hongcheng and Xie, Feiyang and Cai, Penglin and Dong, Hao and Lu, Zongqing},
  journal={arXiv preprint arXiv:2303.16563},
  year={2023}
}

@article{erdogan2025plan,
  title={{PLAN-AND-ACT}: Improving planning of agents for long-horizon tasks},
  author={Erdogan, Lutfi Eren and Lee, Nicholas and Kim, Sehoon and Moon, Suhong and Furuta, Hiroki and Anumanchipalli, Gopala and Keutzer, Kurt and Gholami, Amir},
  journal={arXiv preprint arXiv:2503.09572},
  year={2025}
}

@inproceedings{kannan2024smart,
  title={{Smart-LLM}: Smart multi-agent robot task planning using large language models},
  author={Kannan, Shyam Sundar and Venkatesh, Vishnunandan LN and Min, Byung-Cheol},
  booktitle={2024 IEEE/RSJ International Conference on Intelligent Robots and Systems},
  pages={12140--12147},
  year={2024},
  organization={IEEE}
}

@article{lin2024graph,
  title={Graph-enhanced large language models in asynchronous plan reasoning},
  author={Lin, Fangru and La Malfa, Emanuele and Hofmann, Valentin and Yang, Elle Michelle and Cohn, Anthony and Pierrehumbert, Janet B},
  journal={arXiv preprint arXiv:2402.02805},
  year={2024}
}

@inproceedings{liu2025coherent,
  title={{COHERENT}: Collaboration of heterogeneous multi-robot system with large language models},
  author={Liu, Kehui and Tang, Zixin and Wang, Dong and Wang, Zhigang and Li, Xuelong and Zhao, Bin},
  booktitle={International Conference on Robotics and Automation},
  pages={10208--10214},
  year={2025},
  organization={IEEE}
}

@article{huang2025compositional,
  title={Compositional Coordination for Multi-Robot Teams with Large Language Models},
  author={Huang, Zhehui and Shi, Guangyao and Wu, Yuwei and Kumar, Vijay and Sukhatme, Gaurav S},
  journal={arXiv preprint arXiv:2507.16068},
  year={2025}
}

@inproceedings{yanfangzhou2025m2pa,
  title={{M2PA}: A Multi-Memory Planning Agent for Open Worlds Inspired by Cognitive Theory},
  author={YanfangZhou, YanfangZhou and Li, Xiaodong and Liu, Yuntao and Zhao, Yongqiang and Wang, Xintong and Li, Zhenyu and Tian, Jinlong and Xu, Xinhai},
  booktitle={Findings of the Association for Computational Linguistics: ACL 2025},
  pages={23204--23220},
  year={2025}
}

@article{kagaya2024rap,
  title={{RAP}: Retrieval-augmented planning with contextual memory for multimodal {LLM} agents},
  author={Kagaya, Tomoyuki and Yuan, Thong Jing and Lou, Yuxuan and Karlekar, Jayashree and Pranata, Sugiri and Kinose, Akira and Oguri, Koki and Wick, Felix and You, Yang},
  journal={arXiv preprint arXiv:2402.03610},
  year={2024}
}

@inproceedings{anwar2025remembr,
  title={{REMEMBER}: Building and reasoning over long-horizon spatio-temporal memory for robot navigation},
  author={Anwar, Abrar and Welsh, John and Biswas, Joydeep and Pouya, Soha and Chang, Yan},
  booktitle={International Conference on Robotics and Automation},
  pages={2838--2845},
  year={2025},
  organization={IEEE}
}

@article{li2024optimus,
  title={{OPTIMUS}-1: Hybrid multimodal memory empowered agents excel in long-horizon tasks},
  author={Li, Zaijing and Xie, Yuquan and Shao, Rui and Chen, Gongwei and Jiang, Dongmei and Nie, Liqiang},
  journal={Advances in neural information processing systems},
  volume={37},
  pages={49881--49913},
  year={2024}
}

@article{gupta2026scale,
  title={{Scale-Plan:} {Scalable} Language-Enabled Task Planning for Heterogeneous Multi-Robot Teams},
  author={Gupta, Piyush and Bae, Sangjae and Li, Jiachen and Isele, David},
  journal={arXiv preprint arXiv:2603.08814},
  year={2026}
}

@inproceedings{wang2025karma,
  title={{KARMA}: Augmenting embodied ai agents with long-and-short term memory systems},
  author={Wang, Zixuan and Yu, Bo and Zhao, Junzhe and Sun, Wenhao and Hou, Sai and Liang, Shuai and Hu, Xing and Han, Yinhe and Gan, Yiming},
  booktitle={International Conference on Robotics and Automation},
  pages={1--8},
  year={2025},
  organization={IEEE}
}

@article{agarwal2025l3m+,
  title={{L3M+ P}: Lifelong Planning with Large Language Models},
  author={Agarwal, Krish and Jiang, Yuqian and Hu, Jiaheng and Liu, Bo and Stone, Peter},
  journal={arXiv preprint arXiv:2508.01917},
  year={2025}
}

@inproceedings{hvr_icml_2025,
    title={Hierarchical Planning for Complex Tasks with Knowledge Graph-RAG and Symbolic Verification},
    author={Flavio Petruzzellis and Cristina Cornelio and Pietro Lio},
    booktitle={Forty-Second International Conference on Machine Learning (ICML)},
    year={2025},
}

@inproceedings{
rana2023sayplan,
title={{SayPlan:} Grounding Large Language Models using 3D Scene Graphs for Scalable Task Planning},
author={Krishan Rana and Jesse Haviland and Sourav Garg and Jad Abou-Chakra and Ian Reid and Niko Suenderhauf},
booktitle={7th Annual Conference on Robot Learning},
year={2023},
url={https://openreview.net/forum?id=wMpOMO0Ss7a}
}

@misc{openai2025gpt5,
  title = {{GPT}-5},
  author = {OpenAI},
  year = {2025},
  howpublished = {\url{https://openai.com/gpt-5/}},
  note = {Accessed: 2025-12-03}
}

@article{guan2023leveraging,
  title={Leveraging pre-trained large language models to construct and utilize world models for model-based task planning},
  author={Guan, Lin and Valmeekam, Karthik and Sreedharan, Sarath and Kambhampati, Subbarao},
  journal={Advances in Neural Information Processing Systems},
  volume={36},
  pages={79081--79094},
  year={2023}
}

@misc{llama3herd2024,
  title         = {The {L}lama 3 Herd of Models},
  author        = {{Llama Team, AI @ Meta}},
  year          = {2024},
  eprint        = {2407.21783},
  archivePrefix = {arXiv},
  primaryClass  = {cs.CL},
  url           = {https://arxiv.org/abs/2407.21783}
}

@article{abdin2024phi,
  title={Phi-4 technical report},
  author={Abdin, Marah and Aneja, Jyoti and Behl, Harkirat and Bubeck, S{\'e}bastien and Eldan, Ronen and Gunasekar, Suriya and Harrison, Michael and Hewett, Russell J and Javaheripi, Mojan and Kauffmann, Piero and others},
  journal={arXiv preprint arXiv:2412.08905},
  year={2024}
}

@misc{jiang2023mistral7b,
  title         = {Mistral 7B},
  author        = {Jiang, Albert Q. and Sablayrolles, Alexandre and Roux, Antoine and others},
  year          = {2023},
  eprint        = {2310.06825},
  archivePrefix = {arXiv},
  primaryClass  = {cs.CL},
  url           = {https://arxiv.org/abs/2310.06825}
}

@misc{yang2024qwen2,
  title         = {Qwen2 Technical Report},
  author        = {Yang, An and Yang, Baosong and Hui, Binyuan and others},
  year          = {2024},
  eprint        = {2407.10671},
  archivePrefix = {arXiv},
  primaryClass  = {cs.CL},
  url           = {https://arxiv.org/abs/2407.10671}
}






\end{document}